\documentclass{bmvc2k}
\usepackage{bbding}
\usepackage{pifont}
\usepackage{amssymb}
\usepackage{comment}
\usepackage{multirow}
\usepackage{booktabs}

\usepackage{bm}  
\usepackage{float}
\usepackage{times}
\usepackage{bbding}
\usepackage{xspace}  %
\usepackage{dsfont}
\usepackage{epsfig}
\usepackage{amsmath}
\usepackage{amssymb}
\usepackage{comment}
\usepackage{graphicx}
\usepackage{booktabs}
\usepackage{multirow}
\usepackage{colortbl}
\usepackage{enumitem}
\usepackage{placeins}
\usepackage{footnote}
\usepackage{multicol}
\usepackage{graphbox}
\usepackage{placeins}
\usepackage{mathrsfs} 
\usepackage{enumitem}
\usepackage{makecell}
\usepackage{hyperref}
\usepackage{algorithm}
\usepackage{algorithmic}
\usepackage{threeparttable}

\definecolor{bmvc_blue}{RGB}{0,0,102}
\usepackage{caption}
\title{A Simple Plugin for Transforming Images \\
to Arbitrary Scales}
\addauthor{Qinye Zhou$^*$}{zhouqinye@sjtu.edu.cn}{1}
\addauthor{Ziyi Li$^*$}{599lzy@sjtu.edu.cn}{1}
\addauthor{Weidi Xie$^\dagger$}{weidi@sjtu.edu.cn}{1,2}
\addauthor{Xiaoyun Zhang$^\dagger$}{xiaoyun.zhang@sjtu.edu.cn}{1}
\addauthor{Yanfeng Wang}{wangyanfeng@sjtu.edu.cn}{1,2}
\addauthor{Ya Zhang}{ya_zhang@sjtu.edu.cn}{1,2}

\addinstitution{
 Coop. Medianet Innovation Center,\\
Shanghai Jiao Tong University, China
}
\addinstitution{
Shanghai AI Laboratory
}
\runninghead{ZHOU, ET.AL.}{A Plugin for Transforming Images to Arbitrary Scales}

\def\etal{\emph{et al}\bmvaOneDot}

\begin{document}

\maketitle
\def\thefootnote{*}\footnotetext{Both the authors have contributed equally to this project.  $\dagger$ denote corresponding authors.}\def\thefootnote{\arabic{footnote}}

\begin{abstract}
Existing models on super-resolution often specialized for one scale, 
fundamentally limiting their use in practical scenarios.
In this paper, we aim to develop a general {\em plugin} that can be inserted into existing super-resolution models, conveniently augmenting their ability towards \textbf{A}rbitrary \textbf{R}esolution \textbf{I}mage \textbf{S}caling, thus termed  \textbf{ARIS}.
We make the following contributions:
(i) we propose a transformer-based plugin module, which uses spatial coordinates as query, 
iteratively attend the low-resolution image feature through cross-attention, and output visual feature for the queried spatial location, 
resembling an implicit representation for images;
(ii) we introduce a novel self-supervised training scheme, that exploits consistency constraints to effectively augment the model's ability for upsampling images towards unseen scales, {\em i.e.}~ground-truth high-resolution images are not available;
(iii) without loss of generality, we inject the proposed \textbf{ARIS} plugin module into several existing models, namely, IPT, SwinIR, and HAT, showing that the resulting models can not only maintain their original performance on fixed scale factor but also extrapolate to unseen scales, substantially outperforming existing any-scale super-resolution models on standard benchmarks, {\em e.g.}~Urban100, DIV2K, etc. Project page: \href{https://lipurple.github.io/ARIS_Webpage/}{https://lipurple.github.io/ARIS\_Webpage/}

\end{abstract}

\section{Introduction}

Image super-resolution~(SR) aims to reconstruct high-resolution~(HR) images from corresponding degraded low-resolution~(LR) images. 
In the literature, existing research~\cite{dong2015image,dong2016accelerating,dai2019second,zhang2018residual,wang2018esrgan,kim2016deeply,lu2021efficient,wang2020deep} predominately focuses on training specific models that work well for a few scaling factors, 
thus, different models have to be trained for different factors,
limiting their practical use in real-world applications,
when one may want to scale the image into arbitrary-resolution for viewing purpose.
To address this limitation, some recent approaches, {\em e.g.}~MetaSR~\cite{hu2019meta}, LIIF~\cite{chen2021learning} and LTE~\cite{lee2022local} have considered designing specific architectures for arbitrary-scale super-resolution with a single model. 
Despite being promising, these models still fall behind the existing SR models on 
low-scale super-resolution, 
as we have experimentally shown in Table~\ref{tab:results4}.


In this paper, our goal is to develop a general \textbf{plugin} module that can be inserted into any existing SR models, conveniently augmenting their ability to \textbf{A}rbitrary \textbf{R}esolution \textbf{I}mage \textbf{S}caling, 
thus termed \textbf{ARIS}. 
Specifically, we adopt a transformer-based architecture, 
with spatial coordinates naturally treated as the queries, 
that iteratively attend visual feature of the low-resolution image through an attention mechanism, 
and output the visual representation for desired high-resolution image, 
to be decoded into RGB intensity value at last.
We can continuously scale the image to arbitrary resolution by simply changing the granularity of the spatial coordinates, resembling an implicit representation of the images.

In contrast to LIIF~\cite{chen2021learning} and LTE~\cite{lee2022local}, which also represent an image as a continuous function by MLPs that maps coordinates and the corresponding local latent codes to RGB values, thus achieving arbitrary-scale super-resolution,
our proposed idea poses two critical differences: 
i) we represent the image continuously at the feature level, 
mapping the low-resolution image feature into the high-resolution image feature. 
Thus our module can be inserted into any network without replacing other components and the pre-trained parameters can be re-used directly, 
while LIIF~\cite{chen2021learning} and LTE~\cite{lee2022local} need to retrain the whole network; 
(ii) we use spatial coordinates as query, iteratively attend  the low-resolution image feature through cross-attention, and make full use of the global dependency in images, 
for example, self-similarity, while LIIF~\cite{chen2021learning} and LTE~\cite{lee2022local} take the local latent code as input and have a limited receptive field.

Additionally, for the arbitrary-scale super-resolution task, 
it is often impractical to collect the paired LR-HR images for each scale with high quality, which prevents model from training towards unseen scales. 
To this end, we formulate this problem of reconstruction with incomplete measurements
and introduce a self-supervised training scheme, that scales the image to a target resolution in the absence of paired data, by exploiting consistency constraints.
Specifically, for the scales whose high-resolution images are not available, 
the model is trained by either upsampling the HR images of seen scales or downsampling the reconstructed images towards seen scales. 
As a result, we show this self-supervised training scheme can significantly improve the performance on unseen scales.

To summarise, we consider the problem of arbitrary-scale image super-resolution, and make the following contributions: 
(i) we propose a transformer-based plugin module, called ARIS, 
which resembles an implicit representation for images and can be inserted into any existing super-resolution models, conveniently augmenting their ability to upsample the image with arbitrary scale; 
(ii) we introduce a novel self-supervised training scheme, 
that exploits consistency constraints to train our ARIS plugin module towards out-of-distribution scales, {\em i.e.}, LR-HR image pairs are unavailable;
(iii) 
the ARIS plugin module is orthogonal to the development of new super-resolution architectures,
we insert it into several strong models published recently,
namely, IPT~\cite{chen2021pre}, SwinIR~\cite{liang2021swinir}, HAT~\cite{chen2022activating}, the resulting models outperform the any-scale super-resolution models on various benchmarks, {\em e.g.}~Urban100, DIV2K, {\em etc}.
\section{Related Work}


\paragraph{Image Super-resolution.}
Image super-resolution~(SR) is probably one of the most widely researched problems in  computer vision history, reviewing all the work would be prohibitively impossible, 
we thus only discuss some of the most relevant work.
Early super-resolution approaches are exemplar~\cite{chang2004super,huang2015single} or dictionary~\cite{yang2010image,wang2012semi} based super-resolution. 
These methods generate high-resolution images by using the similarities within and between images.  And the performance is limited by the size of the dataset. 
Since SRCNN~\cite{dong2015image}, 
ConvNets were adopted for solving the image super-resolution task, 
afterwards, numerous deep learning based methods~\cite{dong2015image,dong2016accelerating,dai2019second,kim2016accurate,lai2017deep,lim2017enhanced,zhang2018residual,zhang2018image,wang2018esrgan,kim2016deeply,lu2021efficient,wang2020deep} have been proposed to improve the image quality. 
Specifically, some works innovate the architectural design of the ConvNets, 
for example, the residual block~\cite{lim2017enhanced,zhang2018residual,zhang2018image}, skip-connection~\cite{kim2016accurate,kim2016deeply,mao2016image} and recursive network~\cite{kim2016deeply,tai2017image};
Other works explore different training objectives, 
for example, using adversarial learning~\cite{wang2018esrgan,ledig2017photo}. 
Recently, a series of Transformer-based approaches~\cite{chen2021pre,chen2022activating,liang2021swinir,lu2021efficient} have shown superior performance. 
Generally speaking, these models are often specialised for single-scale super-resolution, 
which fundamentally limits their use in practical scenarios, 
where we may want to scale the image to arbitrary scales. \\[-0.8cm]


\paragraph{Arbitrary-scale  Super-resolution.}
To overcome the above limitation, 
MDSR~\cite{lim2017enhanced}  proposed to integrate a collection of modules that are trained for different scale factors ({\em i.e.}~x2, x3, x4). 
Hu~\etal~\cite{hu2019meta} proposed MetaSR to solve the arbitrary-scale upsampling problem with meta-learning, which directly predicts the filter weights for different scale factors. 
Building on MetaSR, ArbSR~\cite{wang2021learning} designs a scale-aware convolution layer to make better use of the scale information and can handle the problem of asymmetric SR. 
Recently, LIIF~\cite{chen2021learning} introduces the idea of implicit neural representation for images, which treats images as a function of coordinates, thus allowing to scale the image at continuous scales by simply manipulating the spatial grid of the image.
LTE~\cite{lee2022local} introduces a dominant-frequency estimator to allow an implicit function to learn fine details while restoring images in arbitrary resolution. In this paper, we continue the vein of research on using implicit neural representation for image super-resolution, we adopt the transformer-based architecture, 
where the spatial grid can be used as the query in the transformer decoder, 
allowing to iteratively attend the low-resolution image feature through cross-attention.\\[-0.8cm]

\paragraph{Implicit Neural Representation.}
Recent work has demonstrated the great potential of using neural networks as a continuous representation of the signals, for example, for shapes~\cite{genova2019learning,genova2019deep,atzmon2020sal},
objects~\cite{park2019deepsdf,atzmon2020sal,gropp2020implicit}, or scenes~\cite{sitzmann2019scene,jiang2020local,mildenhall2020nerf}. 
Theoretically, such continuous parameterization enables to represent the signal to any level of fine details, with significantly less memory than using a discrete lookup table.
In these representations, an object or scene is usually represented as a multilayer perceptron that maps coordinates to signed distance~\cite{park2019deepsdf,atzmon2020sal,jiang2020local}, occupancy~\cite{mescheder2019occupancy,chen2019learning,niemeyer2020differentiable} or RGB values~\cite{sitzmann2019scene,mildenhall2020nerf}.
In this paper, we focus on learning implicit image representation at the feature level by a transformer-based architecture.

\section{Methods}

\begin{figure}[t]
    \centering
    \includegraphics[width=0.95\linewidth]{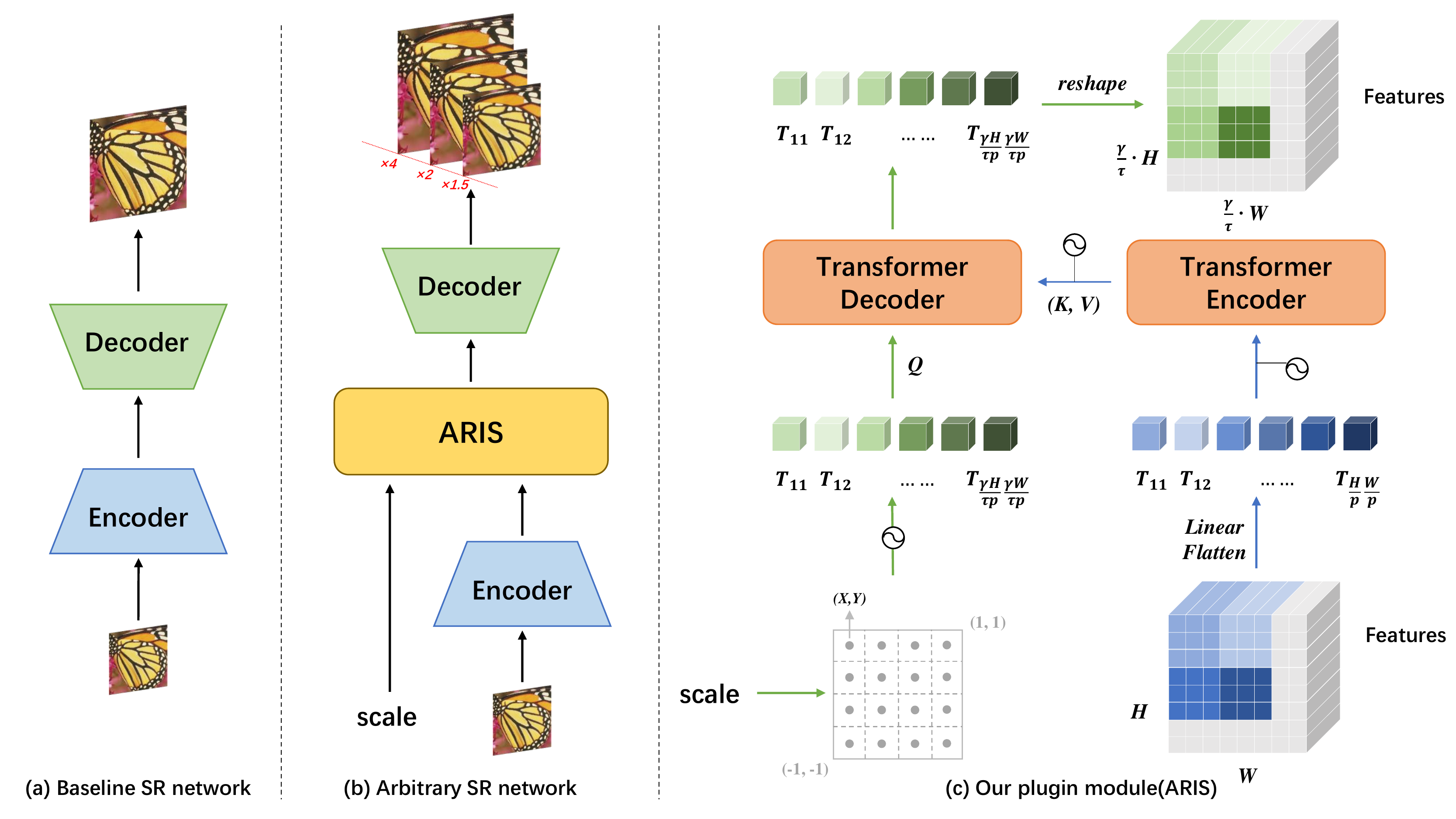}
    \caption{\textbf{An overview of our ARIS plugin module.} Our ARIS plugin module can be inserted into baseline SR network (a) to obtain arbitrary-scale SR network (b). We show the details of the  ARIS plugin module in (c). The ARIS module utilizes the coordinate map (regarded as QUERY) and low-resolution image feature as input and outputs the desired super-resolution image feature.
    }
    \label{fig:framework}
    \vspace{-0.5cm}
\end{figure}

In this paper, our goal is to develop a general plugin module for any existing super-resolution (SR) model that can augment its ability to arbitrary resolution image scaling~(ARIS). The baseline SR network, which refers to the pre-trained scale-specific SR network can be simplified as an encoder-decoder network, 
where the encoder extracts the feature map for the low-resolution input image and the decoder outputs the super-resolution image as shown in Figure \ref{fig:framework}(a). We can obtain the arbitrary-scale SR network by inserting our ARIS plugin module into the baseline SR network as shown in Figure~\ref{fig:framework}(b). \\[-0.5cm]

\paragraph{Overview.}
Assuming we have a training set with $N$ paired low- and high-resolution images, 
$\mathcal{D}_{\text{train}} = \{(\mathcal{X}_{\textsc{LR}}, \mathcal{X}_{\textsc{HR}}^2, \allowbreak \mathcal{X}_{\textsc{HR}}^3, \mathcal{X}_{\textsc{HR}}^4)_n, n \in [1, N]\}$,
where $\mathcal{X}_{\textsc{LR}} \in \mathbb{R}^{H \times W \times 3}$ refers to the low-resolution image, and $\mathcal{X}_{\textsc{HR}}^k \in \mathbb{R}^{kH \times kW \times 3}, \forall k \in [2,3,4]$ refers to its $k\times$ upsampled high-resolution image. The goal is thus to obtain a model that can transform a low-resolution image into arbitrary scales: 

\begin{equation}
    \mathcal{Y}_{\textsc{SR}}^\gamma =\Phi(\mathcal{X}_{\textsc{LR}}, \gamma\text{ })
    = \Phi_{\textsc{Dec}}(\Phi_{\textsc{ARIS}}(\Phi_{\textsc{Enc}}(\mathcal{X}_{\textsc{LR}}), \gamma))
\end{equation}
where $\Phi(\cdot)$ denotes the  trainable function, parameterized by the encoder ($\Phi_{\textsc{E}}$), decoder ($\Phi_{\textsc{D}}$) of the baseline SR network and  our ARIS plugin module ($\Phi_{\textsc{ARIS}}$), that maps a low-resolution image~($\mathcal{X}_{\textsc{LR}}$), 
to the desired super-resolution image~($\mathcal{Y}_{\textsc{SR}}^\gamma \in \mathbb{R}^{\gamma H \times \gamma W \times 3}$), with the scale $\gamma$ denoting continuous values, 
{\em e.g.}~$\gamma \in [1, 8]$. 


In the following sections, 
we first describe the details of the proposed ARIS plugin module
in Section \ref{section3.1}; we then introduce a novel training regime that allows training the model for arbitrary resolution scaling, even without LR-HR image pairs in Section~\ref{section3.3}.

\subsection{ARIS Plugin Module}
\label{section3.1}
Unlike conventional representation that regards an image as a look-up table of intensity values, we use an implicit representation that treats an image as a function mapping from spatial coordinates to intensity, thus allows to continuously scale the image to arbitrary resolution by simply changing the granularity of its spatial coordinates. 
Specifically, we adopt a variant of transformer architecture for our ARIS plugin, with the normalised spatial coordinates as \texttt{query}, 
iteratively attending the visual feature of the low-resolution image, 
to aggregate both local and global information, 
and eventually decode to the image of desired resolution.

As shown in Figure \ref{fig:framework}(c), 
the  ARIS plugin module has two components, 
consisting of transformer encoder and transformer decoder  respectively.
Specifically, the transformer encoder aims to globally aggregate the local visual features from low-resolution images, while the transformer decoder resembles the implicit representation for image, mapping coordinates to visual features for decoding later.

\subsubsection{Transformer Encoder}
Given the visual feature from  the encoder of a baseline SR network, {\em i.e.}~$\Phi_{\textsc{Enc}}(\mathcal{X}_{\textsc{LR}})$,
we use transformer encoder with $L$ layers to aggregate information globally:

\begin{equation}
    \mathcal{F}_{\textsc{LR}}=\Phi_{\textsc{Transformer-E}}(\Phi_{\textsc{Enc}}(\mathcal{X}_{\textsc{LR}}) + \textsc{PE}),
\end{equation}
where  $\Phi_{\textsc{Transformer-E}}(\cdot)$ refers to the transformer encoder.
To maintain the spatial information, 
learnable position encodings~(PE) are added to the  visual features,
and then passed into the transformer as a sequence of tokens. 
As a consequence, features computed from the transformer encoder is denoted as $\mathcal{F}_{\text{LR}} \in \mathbb{R}^{\frac{HW}{p^2} \times C}$,
with $p, C$ referring to the patch size used to generate tokens, 
and feature channels respectively.


\vspace{-0.15cm}

\subsubsection{Transformer Decoder for Implicit Image Representation}
Here, we parametrize the image as a mapping from image coordinates to visual features, 
by adopting a module with multiple transformer decoder layers.
In detail, 
we start by constructing a normalised spatial grid based on the desired scaling factor, 
and project them into high-dimensional vectors with the Fourier encoding~\cite{xiangli2021citynerf},  
$\mathcal{Q}_{\text{SR}} = \Phi_{\textsc{Fourier}}([\mathbf{x}, \mathbf{y}])$,
where $[\cdot, \cdot]$ indicates concatenation of spatial coordinates,
$\mathbf{x} = [-1, \alpha-1, 2\alpha-1, \dots, 1]$, 
$\mathbf{y} = [-1, \beta-1, 2\beta-1, \dots, 1]$ refer to the spatial coordinates respectively, with gaps computed as $\alpha = \frac{\gamma H - 1}{2p\tau}$, $\beta = \frac{\gamma W - 1}{2p\tau}$, where $\tau$ refers to the upsampling scale of  baseline SR network.
As a result, $\mathcal{Q}_{\text{SR}} \in \mathbb{R}^{\frac{\gamma H}{p\tau} \times \frac{\gamma W}{p\tau} \times C}$ denotes the Fourier encoded spatial coordinates for the desired super-resolution image.
\textbf{Note that}, as $\gamma$ can be any continuous value,
the granularity of the spatial coordinates can thus be varying accordingly.

Next, we convert the spatial coordinates map into a sequence of vectors
and used it as \texttt{Query} into a stack of transformer decoder layers~($\Phi_{\textsc{Transformer-D}}(\cdot)$),
\begin{align}
\mathcal{F}_{\text{SR}} = \Phi_{\textsc{Transformer-D}}(W^{Q} \cdot \mathcal{Q}_{\text{SR}}, \text{\hspace{3pt}} W^K \cdot \mathcal{F}_{\text{LR}}, \text{\hspace{3pt}} W^V \cdot \mathcal{F}_{\text{LR}})
\end{align}
where $W^K, W^V$ refer to the learnable parameters that project the visual features to 
\texttt{Key} and \texttt{Value}, $\mathcal{F}_{\text{SR}}$ refers to the enriched visual feature map that can be decoded into desired super-resolution image with decoder,
{\em i.e.}, $\mathcal{Y}_{\textsc{SR}}^\gamma = \Phi_{\textsc{Dec}}(\mathcal{F}_{\textsc{SR}})$. \\[-0.8cm]

\paragraph{Discussion.}
To summarise, the transformer-based ARIS plugin module  can generally adapt  to any existing SR models, enabling them to achieve arbitrary resolution image scaling.  ARIS can  globally aggregate the  visual feature extracted by  the baseline SR network using transformer encoder and further map the feature and spatial coordinates to a visual representation of desired  super-resolution image using transformer decoder, similar to implicit representation for images.

\begin{figure}[t]
    \centering
   \includegraphics[width=0.93\linewidth]{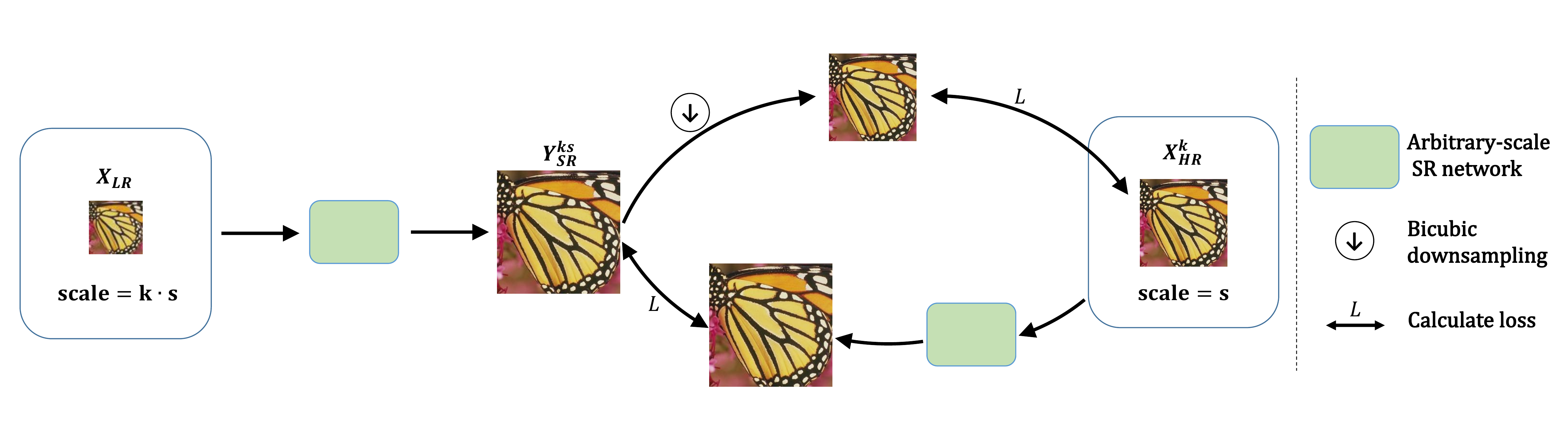}
    \caption{\textbf{Self-supervised training strategy with consistency constraints.} The first training setting is to downsample the SR image ($\mathcal{Y}_{\textsc{SR}}^{ks}$)  to the same resolution as available HR image ($\mathcal{X}_{\textsc{HR}}^{k}$), and thus can supervise, called down-consistency training. The second training setting is to upsample the HR image ($\mathcal{X}_{\textsc{HR}}^{k}$) to the same resolution as the SR image  ($\mathcal{Y}_{\textsc{SR}}^{ks}$) and then supervise using $L1$ loss, called up-consistency training.}
    \label{fig:cycle}
    \vspace{-0.4cm}
\end{figure}

\vspace{-0.15cm}

\subsection{Self-supervised Training Strategy with Consistency Constraints}
\label{section3.3}


It is the common practise in arbitrary scale super-resolution~({\em e.g.}, LIIF~\cite{chen2021learning}), where the scales are divided into in-distribution and out-of-distribution.
In our case, the $\times 2$, $\times 3$ and $\times 4$ are in-distribution~(groundtruth LR-HR pairs are available in the given dataset $\mathcal{D}_{\text{train}}$), $\times 6$ and $\times 8$ are considered as out-of-distribution~(no groundtruth LR-HR pairs). For in-distribution scales, we can train our SR model using the traditional supervision method,
{\em i.e.}, 
$\mathcal L_\text{pair}=L_1( \mathcal{Y}_{\textsc{SR}}^k, \mathcal{X}_{\textsc{HR}}^k)$
where $k \in \{2,3,4\}$ refers to the scale factor, $\mathcal{Y}_{\textsc{SR}}^k=\Phi(\mathcal{X}_{\textsc{LR}}, k)$ refers to the generated super-resolution image. 
For the our-of-distribution scales, 
in order to train the model beyond the resolution limitation, 
{\em i.e.}~the resolution of dataset images might be infeasible for generating LR images for large scales, for example, 
if an HR image is only of resolution $128\times 128$, 
the size of the generated LR image for $\times 8$ is $16\times 16$ at maximum, 
thus it will be infeasible to train the model for  $\times 8$ scaling $32\times 32$ image, we adopt a self-supervised training scheme that exploits consistency constraints.

As shown in Figure \ref{fig:cycle}, 
our proposed training scheme includes two settings, 
{\em i.e.}, down-consistency training and up-consistency training.
Specifically,
we first use our arbitrary-scale SR network to scale the low-resolution image by $k \cdot s$ times,
{\em i.e.}, $\mathcal{Y}_{\textsc{SR}}^{ks}=\Phi(\mathcal{X}_{\textsc{LR}}, k\cdot s)$. For down-consistency training, we downsample the generated super-resolution image to the same resolution as available high-resolution image ($\mathcal{X}_{\textsc{HR}}^{k}$),
and use $L1$ loss as the objective for optimisation.
The down-consistency training method can be formulated as:

\begin{equation}
    \mathcal{L}_{\text{down-consistency}}= |\Phi_{\textsc{Bicubic}}(\mathcal{Y}_{\text{SR}}^{ks}, s) - \mathcal{X}_{\text{HR}}^k|_1
\end{equation}
where $\Phi_{\textsc{Bicubic}}(\cdot, s)$ refers to the simple bicubic downsampling, 
with a factor of $s$.

For up-consistency training,  we use our arbitrary-scale SR network to upsample high-resolution image ($\mathcal{X}_{\textsc{HR}}^{k}$) to the same resolution as the generated super-resolution image ($\mathcal{Y}_{\textsc{SR}}^{ks}$) so that we can supervise. 
It can be formulated as:

\begin{equation}
    \mathcal{L}_{\text{up-consistency}}=|\mathcal{Y}_{\text{SR}}^{ks} -  \Phi(\mathcal{X}_{\textsc{HR}}^k, s)|_1
\end{equation}
Note that, we use $ \mathcal L_\text{pair}$, $\mathcal{L}_{\text{down-consistency}}$ and $\mathcal{L}_{\text{up-consistency}}$ together to train the arbitrary-scale SR network for both seen and unseen scales.


\vspace{-0.15cm}

\section{Experiments}

\subsection{Datasets and Metrics}
We train all models on DIV2K, 
and then evaluate them on five standard benchmark datasets: \\[-0.8cm]

\paragraph{Training Set.} 
DIV2K~\cite{timofte2017ntire} contains over 1000 images in 2K resolution, with 800 images for training, and 100 images for validation and testing, respectively. 
All of our models are trained with DIV2K training set. \\[-0.8cm]

\paragraph{Testing Set.}
Following previous work, we report the performance of our model on 4 benchmark datasets, 
namely, Set5 \cite{bevilacqua2012low}, Set14 \cite{zeyde2010single}, 
B100 \cite{937655}, Urban100 \cite{huang2015single} and the DIV2K validation set.
Note that all the degradation images are generated by the Matlab function {\em imresize} with the default setting of bicubic interpolation. \\[-0.8cm]

\paragraph{Evaluation Metrics.}
In accordance with \cite{chen2021learning,hu2019meta,xu2021ultrasr,chen2021pre,liang2021swinir,lee2022local}, we report peak signal-to-noise ratio (PSNR) on the Y channel of the transformed YCbCr color space for the 4 benchmark datasets and the PSNR on the RGB channel for the DIV2K validation set.

\subsection{Training Details}
\paragraph{Baseline SR Network. } 
In our experiments, we choose IPT~\cite{chen2021pre}, SwinIR\cite{liang2021swinir} and HAT\cite{chen2022activating} as our baseline SR network, 
and only use their pre-trained model on scale ×2. 
As shown in Fig.~\ref{fig:framework}, 
we inject the plugin module between their feature extractor and upsampling layers, 
the resulting models are termed  IPT-ARIS, SwinIR-ARIS, and HAT-ARIS respectively. Each resulting model is trained  individually. Following the design in the original baseline networks, 
we randomly crop $48 \times 48$ patches~
to form the LR images and feed them into the feature extractor and train the model to reconstruct their corresponding HR patches for all models. 
Additionally, we perform data augmentation by randomly rotating 90°, 180°, 270°,
and horizontal flipping. \\[-0.8cm]

\paragraph{Implementation Details.} 
 The setting of training scale factors follows the training strategy described in Section~\ref{section3.3}. Specifically, we use ×2, ×3, ×4, ×6, and ×8 as our training scale factors. For ×2, ×3, and ×4, we have LR-HR pairs, 
 while for ×6 and ×8, we adopt self-supervised training scheme with consistency constraints. 
 For the Transformer Encoder and Transformer Decoder, 
 we both use multi-head attention with 6 heads and 6 layers. 
 For training, we use four NVIDIA Tesla V100 GPUs to train our model for 300 epochs with batch size 8, ADAM~\cite{kingma2014adam} optimiser with $\beta_{1} = 0.9$, $\beta_{2} = 0.999$. 
 The initial learning rate is set to 5e-5 and decayed by one-half at epoch 150, 200, and 250.

\begin{table}[t]
\footnotesize
\begin{center}
\tabcolsep=0.11cm
\begin{tabular}{ccc|cccc|ccc|ccc}
\cmidrule(){1-13}
&  & & \multicolumn{7}{c|}{arbitrary-scale SR method}  & \multicolumn{3}{c}{single-scale SR method} \\
\cmidrule( ){1-13}
Datasets & \multicolumn{1}{c}{scale} & \makecell[c]{Bicubic\\~\cite{lim2017enhanced}} & \makecell[c]{MetaSR\\~\cite{hu2019meta}} & \makecell[c]{ArbSR\\~\cite{wang2021learning}} & \makecell[c]{LIIF\\~\cite{chen2021learning}}  &  \makecell[c]{LTE\\~\cite{lee2022local}} & \makecell[c]{IPT- \\ ARIS} & \makecell[c]{SwinIR-\\ARIS} & \makecell[c]{HAT-\\ ARIS} &  \makecell[c]{IPT\\~\cite{chen2021pre}} & \makecell[c]{SwinIR\\~\cite{liang2021swinir}} & \makecell[c]{HAT\\~\cite{chen2022activating}} \\
\cmidrule( ){1-13}
\cmidrule( ){1-13}
\multirow{5}{*}{Set5}     & ×2 & 33.97  & 38.22 & 38.26  & 38.17 & 38.33 & 38.20    & 38.25       & \textbf{38.50}    &  38.37 & 38.35 & 38.63    \\
                          & ×3 & 30.63  & 34.63 & 34.75  & 34.68 & 34.89 & 34.69    & 34.82       & \textbf{35.00}    &  34.81 & 34.89 & 35.06    \\
                          & ×4 & 28.63  & 32.38 & 32.50  & 32.50 & 32.81 & 32.58    & 32.66       & \textbf{32.94}    &  32.64 & 32.72 & 33.04    \\
                          & ×6 & 26.09  & 29.04 & 28.45  & 29.15 & 29.50 & 29.09    & 29.38       & \textbf{29.63}    &  - & - & -    \\
                          & ×8 & 24.52  & 26.96 & 26.21  & 27.14 & 27.35 & 27.18    & 27.31       & \textbf{27.50}    &  - & - & -    \\
                          \cmidrule( ){1-13}
\multirow{5}{*}{Set14}    & ×2 & 30.55  & 33.98 & 34.07  & 33.97 & 34.25  & 33.94    & 34.21       & \textbf{34.81}    & 34.43 & 34.14 & 34.86   \\
                          & ×3 & 27.79  & 30.54 & 30.64  & 30.53 & 30.80 & 30.64    & 30.75       & \textbf{31.05}    &  30.85 & 30.77 & 31.08    \\
                          & ×4 & 26.21  & 28.78 & 28.84  & 28.80 & 29.06 & 28.92    & 29.01       & \textbf{29.22}    &  29.01 & 28.94 & 29.23    \\
                          & ×6 & 24.44  & 26.51 & 26.22  & 26.64 & 26.86 & 26.61    & 26.79       & \textbf{26.96}    &  - & - & -    \\
                          & ×8 & 23.28  & 24.97 & 24.55  &  25.15 & 25.42 & 25.11    & 25.27       & \textbf{25.47}    &  - & - & -    \\
                          \cmidrule( ){1-13}
\multirow{5}{*}{B100}     & ×2 & 29.73  & 32.33 & 32.38  & 32.32 & 32.44 & 32.35    & 32.42       & \textbf{32.59}    &  32.48 & 32.44 & 32.62    \\
                          & ×3 & 27.31  & 29.26 & 29.31  & 29.26 & 29.39 & 29.30    & 29.36       & \textbf{29.49}    &  29.38 & 29.37 & 29.54    \\
                          & ×4 & 26.04  & 27.71 & 27.74  &  27.74 & 27.86 & 27.78    & 27.84       & \textbf{27.98}    &  27.82 & 27.83 & 28.00    \\
                          & ×6 & 24.61  & 25.90 & 25.74  & 25.98 & 26.09 & 25.97    & 26.04       & \textbf{26.16}    &  - & - & -    \\
                          & ×8 & 23.73  & 24.83 & 24.55  & 24.91 & 25.03 & 24.92    & 24.98       & \textbf{25.09}    &  - & - & -    \\
                          \cmidrule( ){1-13}
\multirow{5}{*}{Urban100} & ×2 & 27.07  & 32.92 & 33.07  & 32.87 & 33.50  & 33.17   & 33.35       & \textbf{34.28}    &  33.76 & 33.40 & 34.45    \\
                          & ×3 & 24.58  & 28.82 & 28.97  & 28.82 & 29.41 & 29.21    & 29.31       & \textbf{30.01}    &  29.49 & 29.29 & 30.23    \\
                          & ×4 & 23.24  & 26.55 & 26.63  & 26.68 & 27.24 & 27.13    & 27.21       & \textbf{27.84}    &  27.26 & 27.07 & 27.97    \\
                          & ×6 & 21.71  & 23.99 & 23.70  & 24.20 & 24.62 & 24.43    & 24.50       & \textbf{25.00}    &  - & - & -    \\
                          & ×8 & 20.80  & 22.59 & 22.13  & 22.79 & 23.17 & 23.16    & 23.08       & \textbf{23.54}    &  - & - & -    \\
                          \cmidrule( ){1-13}
\multirow{5}{*}{DIV2K}    & ×2 & 31.24  & 35.00 & 34.97  & 34.99 & 35.24 & 34.99    & 35.13       & \textbf{35.41}    &  - & - & -    \\
                          & ×3 & 28.37  & 31.27 & 31.28  & 31.26 & 31.50 & 31.32    & 31.44       & \textbf{31.67}    &  - & - & -    \\
                          & ×4 & 26.78  & 29.25 & 29.23  & 29.27 & 29.51 & 29.37    & 29.47       & \textbf{29.70}    &  - & - & -    \\
                          & ×6 & 24.93  & 26.88 & 26.61  & 26.99 & 27.20 & 27.03    & 27.10       & \textbf{27.31}    &  - & - & -    \\
                          & ×8 & 23.78  & 25.57 & 24.99  & 25.61 & 25.81 & 25.64    & 25.71       & \textbf{25.91}    &  - & - & -    \\
                          \cmidrule( ){1-13}
\end{tabular}
\end{center}
\vspace{-5pt}
\caption{\textbf{Quantitative comparison with single-scale SR methods and other state-of-the-art methods for arbitrary-scale super-resolution on five benchmark datasets.} The bold numbers indicate the best results in arbitrary-scale SR methods. Arbitrary-scale SR methods train a single  model for all scales. Single-scale SR methods train a specific model for each scale. }
\label{tab:results4}
\vspace{-0.45cm}
\end{table}


\vspace{-0.2cm}

\section{Results}

As shown in Table.~\ref{tab:results4}, 
we provide experimental results for our proposed plugin injected into different baseline networks,
then compare with other state-of-the-art approaches for arbitrary-scale super-resolution.
After that, we conduct a series of ablation studies on different architectural design choices and training strategies.

\subsection{Quantitative Results}

\paragraph{Comparison to State-of-the-art.}
When comparing to the existing approaches for arbitrary-scale super-resolution, 
due to the architectural difference, we only include the best numbers.
As shown in Table~\ref{tab:results4}, 
adding our proposed ARIS plugin to any baseline SR network can achieve competitive performance on all benchmarks,
and specifically, HAT-ARIS can outperform all existing arbitrary-scale super-resolution models, for both in- and out-distribution scales, validating the effectiveness of ARIS.\\[-0.8cm]

\paragraph{Compare to Baseline SR Network.}
We compare the SR results on baseline networks before and after injecting the ARIS plugin.
The IPT-ARIS, SwinIR-ARIS, and HAT-ARIS achieve comparable performance to their corresponding baseline networks on in-distribution (seen) scale factors~($\times 2$, $\times 3 $ and $\times 4$) and are able to extrapolate to out-distribution scales~($\times 6$, $\times 8 $). Notably, the baseline SR networks are trained for a specific scale, thus they may have more advantages on a specific task than our method. Our plugin module inherits the original ability of baseline SR networks and fits more scales effectively. \\[-0.8cm]

\subsection{Ablation Studies}\label{section5.2}
In this section, we perform ablation studies on the necessity of self-supervised training scheme with consistency constraints,
due to the space limitation, we leave the investigation to the patch size and the training scale in the supplementary material.

\begin{figure}[t]
    \centering
    \includegraphics[width=.98\linewidth]{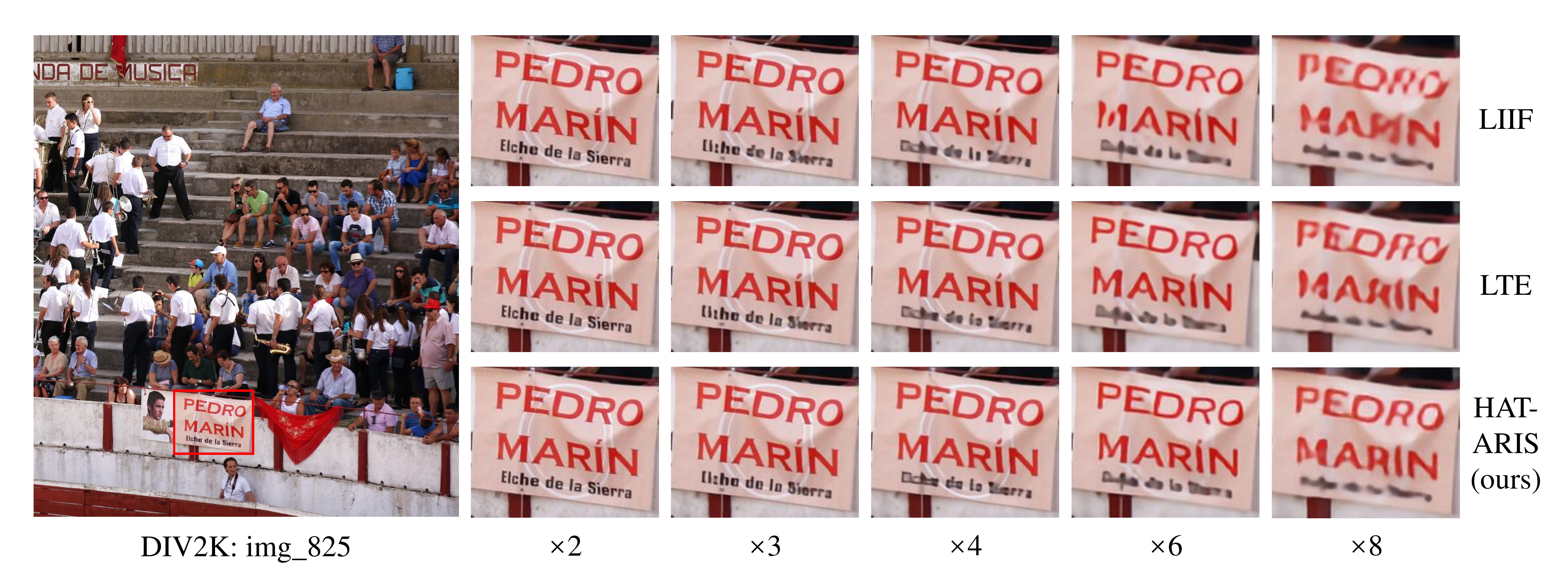}
    \caption{\textbf{Qualitative comparison of the same patch under different upsampling scale factors.} Our method recovers clearer edges of printed texts.}
    \label{fig:result2}
    \vspace{-0.5cm}
\end{figure}

\paragraph{Self-supervised Training Scheme with Consistency Constraints.}
In Table.~\ref{tab:training}, 
we use the standard DIV2K training dataset as our training set, 
which contains paired images in different scaling factors~($\times 2$, $\times 3 $ and $\times 4$). Without self-supervised training, we can achieve good performance on training scales, however, the model performs poorly on the unseen scale factors, {\em e.g.}, ARIS-A. 
While with our proposed self-supervised training,
the models can effectively improve the performance on unseen scales, {\em e.g.}, ARIS-(B, C, D), showing it enables to train SR models even with incomplete measurements to some extent.

\begin{table}[h]
\footnotesize
\begin{center}
\tabcolsep=0.2cm
\begin{tabular}{ccccccccccccc}
\cmidrule( ){1-13}
& \multicolumn{7}{c}{training scale}                                                                                       & \multicolumn{5}{c}{testing scale}                                                                                        \\ \cmidrule(lr){2-8}\cmidrule(lr){9-13}
& \multicolumn{1}{c}{×2}    & \multicolumn{1}{c}{×3}    & \multicolumn{1}{c}{×4}    & \multicolumn{1}{c}{×6 $\downarrow$}    & ×8$\downarrow$ & \multicolumn{1}{c}{×6$\uparrow$}    & ×8$\uparrow$    & \multicolumn{1}{c}{×2}    & \multicolumn{1}{c}{×3}    & \multicolumn{1}{c}{×4}    & \multicolumn{1}{c}{×6}    & ×8    \\ \cmidrule( ){1-13}
ARIS-A & \checkmark  & \checkmark  & \checkmark  &    &   &    &   & \multicolumn{1}{c}{35.48} & \multicolumn{1}{c}{31.72} & \multicolumn{1}{c}{29.72} & \multicolumn{1}{c}{25.57} & \multicolumn{1}{c}{23.84} \\ 
ARIS-B & \checkmark  & \checkmark   & \checkmark  & \checkmark  & \checkmark &    &    & \multicolumn{1}{c}{35.45} & \multicolumn{1}{c}{31.69} & \multicolumn{1}{c}{29.70} & \multicolumn{1}{c}{27.02} & \multicolumn{1}{c}{25.59} \\ 
ARIS-C & \checkmark  & \checkmark   & \checkmark  &    &    & \checkmark  & \checkmark & \multicolumn{1}{c}{35.38} & \multicolumn{1}{c}{31.66} & \multicolumn{1}{c}{29.69} & \multicolumn{1}{c}{\textbf{27.31}} & \multicolumn{1}{c}{25.89} \\ 
\textbf{ARIS-D} & \checkmark  & \checkmark   & \checkmark  & \checkmark  & \checkmark & \checkmark  & \checkmark & \multicolumn{1}{c}{35.41}      & \multicolumn{1}{c}{31.67}      & \multicolumn{1}{c}{29.70}      & \multicolumn{1}{c}{\textbf{27.31}}      &  \multicolumn{1}{c}{\textbf{25.91}}  \\ \cmidrule( ){1-13}
\end{tabular}
\end{center}
\vspace{-10pt}
\caption{\textbf{Ablation study on training strategy.} Performance is measured by PSNR on DIV2K validation set. Specifically, $\downarrow$ and $\uparrow$ denote the down-consistency training and up-consistency  training.}
\label{tab:training}
\vspace{-.5cm}
\end{table}


\vspace{-0.15cm}
\paragraph{Transformer Layers.}
We ablate the number of heads and layers of our Transformer Encoder and Decoder on the Set5 and Set14 dataset with $\times 2$, $\times 3 $ and $\times 4$ scales based on HAT-ARIS in Table~\ref{tab:transformer2}. 
As more layers and heads are added, the performance consistently improves, 
achieving the best performance with 6 layers and 6 heads. 

\vspace{-0.15cm}

\begin{table}[!htbp]

\footnotesize
\begin{center}
\tabcolsep=0.3cm
\begin{tabular}{ccccccccc}
\cmidrule( ){1-9}
\multicolumn{1}{l}{\multirow{2}{*}{layer}} & \multicolumn{1}{l}{\multirow{2}{*}{head}} & \multirow{2}{*}{parameters} & \multicolumn{3}{c}{Set5} & \multicolumn{3}{c}{Set14} \\
\cmidrule(lr){4-6}\cmidrule(lr){7-9}
\multicolumn{1}{l}{}                       & \multicolumn{1}{l}{}                      &                             & ×2     & ×3     & ×4     & ×2      & ×3     & ×4     \\
\cmidrule( ){1-9}
\multirow{2}{*}{3}                         & 3                                         & 200M                        & 38.51  & 34.81  & 32.07  & 34.71   & 30.64  & 28.84  \\
                                           & 6                                         & 200M                        & 38.55  & 34.94  & 32.87  & 34.84   & 31.05  & 29.19  \\
                                           \cmidrule( ){1-9}
\multirow{2}{*}{6}                         & 3                                         & 307M                        & 38.50  & 34.76  & 31.99  & 34.62   & 30.53  & 28.82  \\
                                           & 6                                         & 307M                        & \textbf{38.57}  & \textbf{35.04}  & \textbf{32.98}  & \textbf{34.87}   & \textbf{31.10}  & \textbf{29.23} \\
                                           \cmidrule( ){1-9}
\end{tabular}

\end{center}
\vspace{-0.3cm}
\caption{\textbf{Ablation study on Transformer.} Performance increases with layers and heads.}
\label{tab:transformer2}

\vspace{-0.45cm}

\end{table}




\begin{figure}[!htb]
    \centering
    \includegraphics[width=\linewidth]{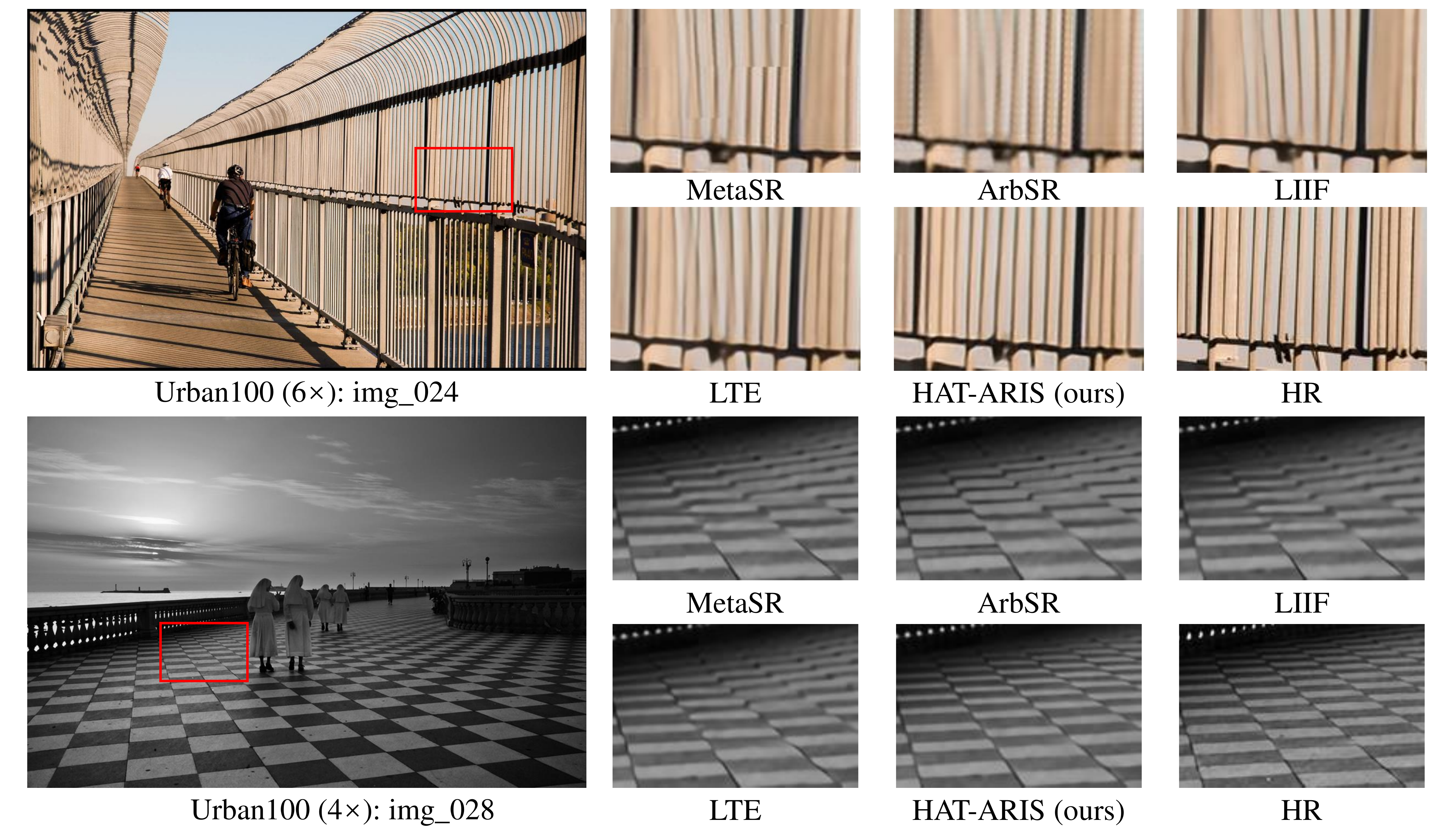}
    \caption{\textbf{Visual results with bicubic downsampling from Urban100}. The patches for comparison are marked with \textcolor{red}{red} boxes in the original images. Our method recovers more details and achieves a better visual effect.}
    \label{fig:result1}
    \vspace{-0.3cm}
\end{figure}

\subsection{Qualitative Results}

We demonstrate the qualitative results in Figure~\ref{fig:result2} and Figure~\ref{fig:result1}, 
 making the following observations: 
$\mathit{First}$, compared with other state-of-the-art methods, the image generated by our model recovers more details and has higher fidelity, while previous methods can not recover the original images and generate some irregular shapes;
$\mathit{Second}$, our model can learn a better continuous image representation. For example, in Figure~\ref{fig:result1}, LIIF and LTE can not restore clear edges of the texts. In contrast, our HAT-ARIS is capable of maintaining the shape information better. This becomes more evident with the increase of scale factors. 




\vspace{-0.2cm}
\section{Limitations}
As ARIS adopts a Transformer architecture, it incurs relatively high memory consumption, and poses poses limitations for extending the model to finer granularity, {\em i.e.}, decimal scale factor with a small stride. 
As future work, we will investigate more efficient transformer architectures, 
for example, using local attention to replace the full attention, 
consider to sample effective tokens while computing attentions~\cite{fayyaz2021ats,yin2022vit}.

\vspace{-0.2cm}
\section{Conclusion}
In this paper, we propose ARIS, a transformer-based plugin module that can be injected into any super-resolution models and augment them towards arbitrary super-resolution. Specifically, we represent the image continuously by using spatial coordinates as query and mapping the low-resolution features into high-resolution features. We introduce a self-supervised training scheme with consistency constraints that can effectively augment the model's ability on unseen scales. Extensive experiments show that the proposed
plugin module outperforms existing state-of-the-art arbitrary-SR methods on five benchmark datasets for all scale factors, showing the great potential of learning image implicit representation. 
\bibliography{egbib}
\end{document}